%% file: main.tex
\documentclass{article}

\PassOptionsToPackage{numbers, compress}{natbib}

\usepackage[final]{neurips_2018}

\usepackage[utf8]{inputenc} 
\usepackage[T1]{fontenc}    
\usepackage{hyperref}       
\usepackage{url}            
\usepackage{booktabs}       
\usepackage{amsfonts}       
\usepackage{nicefrac}       
\usepackage{microtype}      

\usepackage{color}
\usepackage{graphicx}
\usepackage{xspace}
\usepackage{amsmath}
\usepackage{sidecap}  
\usepackage{floatrow}
\usepackage{caption}
\usepackage{subcaption}
\usepackage{epsfig}
\usepackage{multirow}
\usepackage{epsfig}
\usepackage{subcaption}
\usepackage{comment}
\usepackage{amsthm}
\usepackage{gensymb}
\sidecaptionvpos{figure}{c}

\input{math_commands.tex}

\title{Self-Supervised GAN to Counter Forgetting}

\author{
  Ting Chen\thanks{Work done at Google}\, 
  \thanks{Equal contribution} \\
  University of California, Los Angeles \\
  \texttt{tingchen@cs.ucla.edu} \\
  \And
  Xiaohua Zhai\footnotemark[2] \\
  Google Brain \\
  \texttt{xzhai@google.com} \\
  \And
  Neil Houlsby \\
  Google Brain \\
  \texttt{neilhoulsby@google.com} \\
}

\newcommand{\nop}[1]{}

\begin{document}

\maketitle

\input{content/abs}
\input{content/intro}

\input{content/model}
\input{content/exp}

\input{content/related}

\paragraph{Acknowledgments}
We would like to thank Mario Lucic, Marvin Ritter, Sylvain Gelly, Ilya Tolstikhin, Alexander Kolesnikov and Lucas Beyer for help with and discussions on this project.

{\small\bibliographystyle{unsrtnat}
{\small\bibliography{gan}}}

\newpage
\appendix
\input{content/app_model}
\input{content/app_exp}

\end{document}

%% file: math_commands.tex
\usepackage{amsmath,amsfonts,bm}

\def\eqref#1{equation~\ref{#1}}

\def\1{\bm{1}}

\DeclareMathAlphabet{\mathsfit}{\encodingdefault}{\sfdefault}{m}{sl}
\SetMathAlphabet{\mathsfit}{bold}{\encodingdefault}{\sfdefault}{bx}{n}

\DeclareMathOperator{\Tr}{Tr}

\DeclareMathOperator{\FID}{FID} 

\newcommand{\celebahq}{\textsc{celeba-hq}}
\newcommand{\lsun}{\textsc{lsun-bedroom}}
\newcommand{\cifar}{\textsc{cifar10}}
\newcommand{\imagenet}{\textsc{imagenet}}

%% file: content/abs.tex
\begin{abstract}
GANs involve training two networks in an adversarial game, where each network's task depends on its adversary.   
Recently, several works have framed GAN training as an online or continual learning problem~\citep{thanh2018catastrophic,anonymous2019generative,seff2017continual,shrivastava2017learning,kim2018memorization,grnarova2017online}.
We focus on the discriminator, which must perform classification under an (adversarially) shifting data distribution.
When trained on sequential tasks, neural networks exhibit \emph{forgetting}.
For GANs, discriminator forgetting leads to training instability~\citep{thanh2018catastrophic}.
To counter forgetting, we encourage the discriminator to maintain useful representations by adding a self-supervision.
Conditional GANs have a similar effect using labels.
However, our self-supervised GAN does not require labels, and closes the performance gap between conditional and unconditional models.
We show that, in doing so, the self-supervised discriminator learns better representations than regular GANs.
\footnote[1]{Full version of this paper available at \texttt{https://arxiv.org/pdf/1811.11212.pdf}}.
\end{abstract}

%% file: content/intro.tex
\section{Introduction}


Generative Adversarial Networks (GANs) are a class of unsupervised generative models~\citep{goodfellow2014generative}.
GANs involve training a \emph{generator} and \emph{discriminator} network in an adversarial game, such that the generator learns to produce samples from a desired data distribution.
Training GANs is challenging because one searches for a Nash equilibrium of a non-convex game in a high-dimensional parameter space.
GANs are typically trained with alternating stochastic gradient descent.
However, this training procedure is unstable and lacks guarantees~\cite{salimans2016improved}.
Training may exhibit instability, divergence, cyclic behaviour, or mode collapse~\citep{mescheder2018training}.
As a result, many works propose techniques to stabilize GAN training~\citep{mao2016least,gulrajani2017improved,miyato2018spectral,abn,radford2016,zhang2018self,karras2017progressive}.


Recently, several papers frame GAN training as a continual, or online, learning problem~\citep{thanh2018catastrophic,anonymous2019generative,seff2017continual,shrivastava2017learning,kim2018memorization,grnarova2017online}.
GANs may be viewed as online learning because each network, the generator and discriminator, must learn in the context of a non-stationary environment.
The discriminator, for example, is a binary classifier where one class (the fake samples) is non-stationary.
With this view, \citet{thanh2018catastrophic} study catastrophic forgetting in the discriminator and mode collapse, relating these to training instability.
\citet{anonymous2019generative} counter discriminator forgetting by leveraging techniques from continual learning directly (Elastic Weight Sharing~\cite{kirkpatrick2017overcoming} and Intelligent Synapses~\citep{zenke2017continual}).
Other works address forgetting in GANs by retraining on old data, or adding an explicit memory~\citep{shrivastava2017learning,kim2018memorization}.
Finally, \citet{seff2017continual} extend continual learning for GANs by training on nonstationary true-data distributions also.


We follow this line of work, but propose an alternative approach.
Instead of adding explicit replay or memorization strategies, we apply representation learning.
We train the discriminator to learn stable representations that are useful for identifying real images. 
We propose to add self-supervision to the discriminator, 
in particular, we apply a rotation-based loss~\citep{gidaris2018unsupervised}.

Conditional GANs, which use side information such as class labels, are state-of-the-art in generating high fidelity complex images~\cite{mirza2014conditional,odena2017,miyato2018cgans}.
A major contributor to their success is augmentation of the discriminator with supervised information, which encourages it to learn stable and diverse representations.
However, labels are not always available, and unconditional GANs perform much worse.
Our self-supervised GAN requires no additional information, and closes the performance gap for image generation between conditional and unconditional GANs.
As intended, we observe that the self-supervised GAN learns better discriminator representations than other GANs for the task of image classification.
We hope that this work encourages further investigation into representation and continual learning to improve GANs.

%% file: content/model.tex
\section{The Self-supervised GAN}

We first discuss discriminator forgetting and motivate representation learning for GANs. We then present our solution using rotation based self-supervision.

\subsection{Discriminator Forgetting}

The original value function for GAN training is:
\begin{equation}
\label{eq:gan_obj_classic}
V(G, D) = \mathbb{E}_{\bm x\sim P_{\mathrm{data}}(\bm x)}[\log P_D(S=\mathrm{real}|\bm x)] 
+ \mathbb{E}_{\bm x\sim P_{G}(\bm x)}[\log (1- P_D(S=\mathrm{real}|\bm x))]
\end{equation}
where $P_{\mathrm{data}}$ is the true data distribution, and $P_G$ is the data distribution
induced by feeding noise drawn from a simple distribution $\bm z\sim P(\bm z)$ through the generator, $\bm x = G(\bm z)$.
$P_D(S|\bm x)$ is the discriminator's Bernoulli distribution over the sources (real or fake).


The generator maximizes Equation~\ref{eq:gan_obj_classic}, while the discriminator minimizes it. 
Training is typically performed via alternating stochastic gradient descent.
The discriminator, therefore, classifies between two data distributions, $P_{\mathrm{data}}(\bm x)$, and $P_G^{(t)}(\bm x)$, where $t$ indexes training iterations.
The latter, however, is non-stationary because the generator's parameters are updated over time.
This is an online learning problem for which explicit temporal dependencies have been proposed to improve training~\citep{anonymous2019generative,shrivastava2017learning,grnarova2017online,salimans2016improved}.

\begin{figure}[ht]
\begin{center}
\epsfig{file=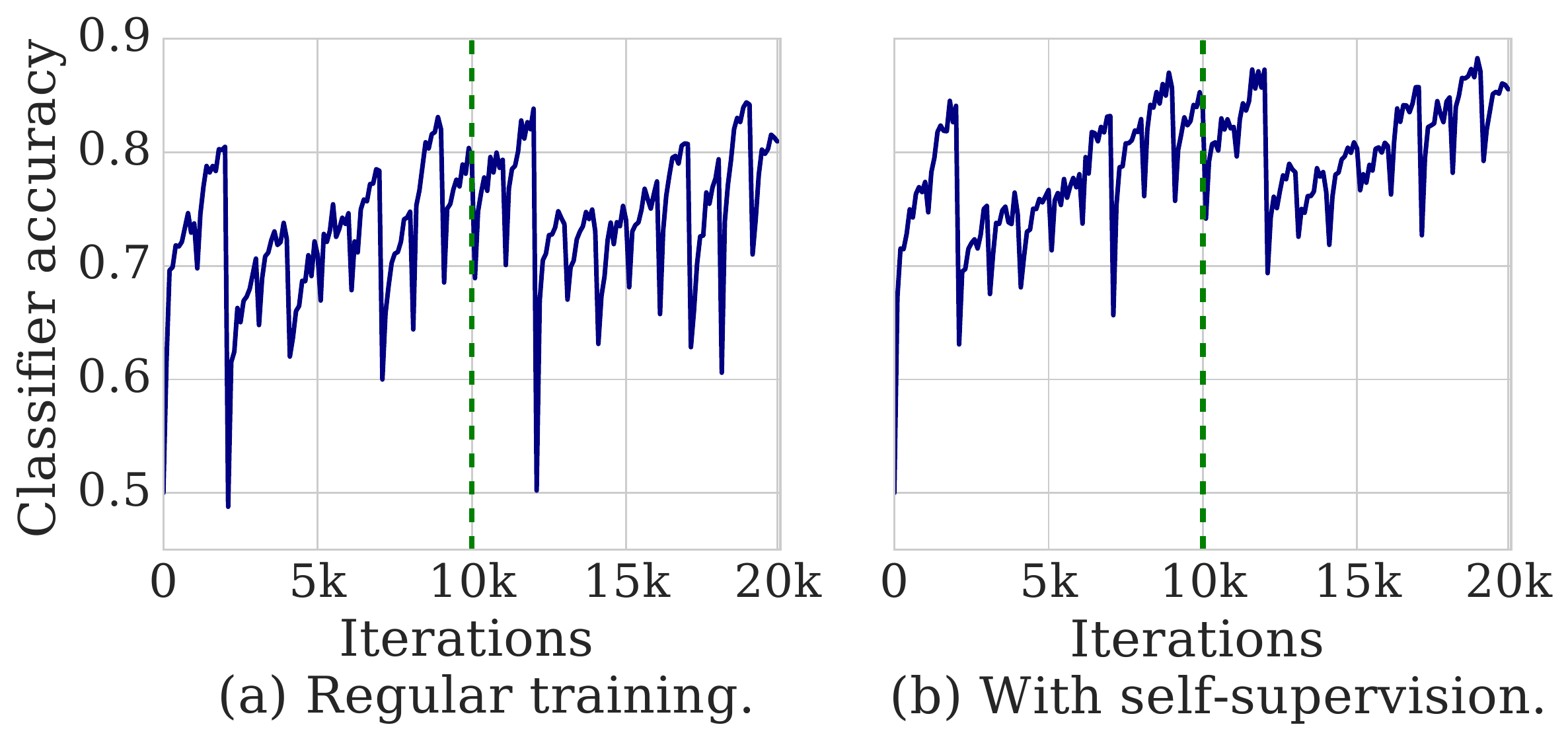,height=4.5cm}
\end{center}
\caption{\label{fig:non-stationary} 
Image classification accuracy when the underlying class distribution shifts every $1$k iterations.
The vertical dashed line indicates the end of an entire cycle, and return to the original classification task at $t=0$.
\emph{Left}: vanilla classifier.
\emph{Right}: classifier with additional self-supervised loss.
This example demonstrates that a classifier/discriminator may fail to learn generalizable representations in a non-stationary environment, but an auxiliary loss mitigates this problem.}
\end{figure}

In online learning, neural networks tend to forget previous tasks~\cite{kirkpatrick2017overcoming,mccloskey1989catastrophic,french1999catastrophic}.
We demonstrate forgetting in a toy setting, Figure~\ref{fig:non-stationary}(a).
Here, a classifier to performs sequential 1-vs.-all classification task on each of the ten classes in \textsc{cifar10}. 
The classifier is trained for $1$k iterations on each task before switching to the next.
Figure~\ref{fig:non-stationary}(a) shows substantial forgetting, despite the tasks being similar.
The classifier returns almost to random accuracy, $0.5$, each time the task switches.
The accuracy even drops substantially when returning to the original task at iteration $10$k.
This demonstrates that the model does not retain generalizable representations during online learning.

Consideration of convergence provides further indication that discriminators may forget.
\citet{goodfellow2014generative} show that the optimal discriminator estimates the likelihood ratio between the generated and real data distributions. 
Therefore, given a perfect generator, where $P_G = P_D$, the optimal discriminator simply outputs $0.5$, and has no requirement to retain any meaningful representations.

Discriminator forgetting may cause training difficulties because 
(1) it does not learn meaningful representations to guide the generator, and 
(2) the generator can revert to generating old images to fool it~\citep{thanh2018catastrophic}.
This is particularly problematic if the data is diverse and complex and the generator exhibits mode collapse.
Therefore, we add self-supervision to encourage the discriminator to retain useful representations.

\subsection{The Self-Supervised GAN}

Self-supervised learning is a family of methods for building representations from unsupervised data.
Self-supervision works by creating artificial supervised tasks from unsupervised data, training on these tasks, and then extracting representations from the resulting networks~\citep{dosovitskiy2014,doersch2015unsupervised,zhang2016colorful}.
Here, we apply the successful image rotation self-supervision method~\citep{gidaris2018unsupervised}.
In this method, the self-supervised network predicts the angle of rotation of an image.
In Figure~\ref{fig:non-stationary}(b) we motivate this loss using our toy problem.
When we add the self-supervised loss, the network learns features that transfer across tasks; performance continually improves, and does not drop to $0.5$ when the distribution shifts.

For the self-supervised GAN, the specific losses we use for the generator and discriminator are:
\begin{align}
L_G &= -V(G, D) -\alpha \mathbb{E}_{\bm x\sim P_G}[\sum_{r\in \mathcal{R}}\nolimits \log P^\prime_{D}(R=r|{\bm x}^r)],\\
L_D &= V(G, D) -\beta \mathbb{E}_{\bm x \sim P_{\mathrm{data}}}[\sum_{r\in \mathcal{R}}\nolimits \log P^\prime_{D}(R=r|{\bm x}^r)],
\end{align}
where $V(G, D)$ is the original GAN loss in Equation~\ref{eq:gan_obj_classic}.
$r\in\mathcal{R}$ is a rotation selected from a set of possible rotations.
We use 
$\{ 0^{\degree}, 90^{\degree}, 180^{\degree}, 270^{\degree} \}$
as in~\citet{gidaris2018unsupervised}.
$P'_D(R|\bm x^r)$ is the discriminator's distribution over rotations,
and $\bm x^r$ is the image $\bm x$ transformed by rotation $r$.
Architecturally, we use a single discriminator network with two heads to
compute $P_D$ and $P^\prime_{D}$. See supplementary Figure~\ref{fig:rotation_acgan} for an overview.

\paragraph{A note on convergence}
With $\alpha>0$ convergence, even under optimal conditions, to the true data distribution $P_G=P_{\mathrm{data}}$ is not guaranteed.
This may not be a concern because current GANs are far from attaining the optimal solution.
If it is, one could anneal $\alpha$ to zero during training.
Our intuition is that the proposed loss encourages the discriminator to learn and retain meaningful representations that allow it to distinguish rotations as well as true/fake images.
The generator is then trained to match distributions in this feature space which encourages the generation of realistic objects.

%% file: content/exp.tex
\section{Experiments}
\label{sec:experiments}

We show that self-supervision improves GAN training, even matching the performance of conditional GANs. 
We also demonstrate, by way of representation evaluation, that the features learned by the self-supervised GAN are more meaningful than those of a vanilla GAN.

\paragraph{Datasets and Settings}
Our main result uses the \textsc{imagenet} dataset.
For the self-supervised GAN (SsGAN), we adopt the architectures of, and compare to, the unconditional GAN in~\citet{miyato2018spectral} (SN-GAN) and conditional GAN in~\citet{miyato2018cgans} (P-cGAN).
Both the generator and discriminator consist of ResNet architectures.
For the conditional generator in P-cGAN, we apply label-conditional batch norm.
SsGAN does not use labels, so by default the generator does not use condition batch-norm, as in SN-GAN.
Therefore, we also consider a variant of the SsGAN with self-modulated batch norm (sBN)~\citep{abn} on the generator to replace label-conditional batch norm. 
We compare image generation quality using the Frechet Inception Distance (FID)~\citep{heusel2017gans}.
We train the model for 1M steps using a single P100 GPU.
Further details and results can be found in the supplementary material.

We also test three smaller datasets, \textsc{cifar10}, \textsc{lsun-bedroon}, and \textsc{celeba-hq}.
Labels for Pc-GAN are only available for \textsc{cifar10}.
For all datasets we set $\beta=1$ and
$\alpha=0.2$, chosen using a small hyperparameter sweep.
For all other hyperparameters, we use the values in \citet{miyato2018spectral} and \citet{miyato2018cgans}.
We train on these datasets for $100$k steps on a single P100 GPU.

\paragraph{Image generation}
Figure~\ref{fig:convergence_fid_imagenet_1m} shows FID training curves on 
\textsc{imagenet}.
The unconditional GAN is unstable, and sometimes diverges.
The conditional P-cGAN substantially outperforms the unconditional SN-GAN.
The SsGAN is stable, and even performs as well as the P-cGAN.

\begin{minipage}{\textwidth}
\centering
\begin{minipage}[b]{0.53\textwidth}
    \centering
    \includegraphics[width=\linewidth]{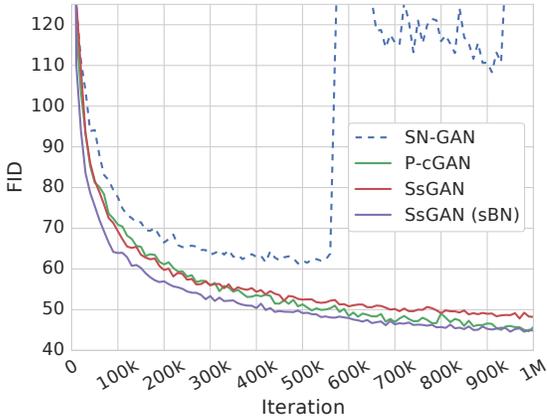}
    \captionof{figure}{\label{fig:convergence_fid_imagenet_1m}Mean training curves across three random seeds.}
\end{minipage}
\hfill
\begin{minipage}[b]{0.42\textwidth}
\vspace{0pt}
    \centering
    \small
    \begin{tabular}{lll}
    \toprule
    Dataset & Method &   FID   \\
    \midrule
    \multirow{4}{*}{\textsc{cifar10}} & SN-GAN &     19.725 \\
            & P-cGAN &     \textbf{15.604} \\
            & SsGAN &     17.114 \\
            & SsGAN (sBN) &     \textbf{15.653} \\ \midrule
    \multirow{4}{*}{IMAGENET} & SN-GAN &     56.670 \\
            & P-cGAN &     \textbf{42.071} \\
            & SsGAN &     47.555 \\
            & SsGAN (sBN) &     43.869 \\ \midrule
    \multirow{3}{*}{BEDROOM} & SN-GAN &     16.021 \\
            & SsGAN &     13.664 \\
            & SsGAN (sBN) &     \textbf{13.298} \\ \midrule
    \multirow{3}{*}{CELEBA-HQ} & SN-GAN &     \textbf{23.765} \\
            & SsGAN &     26.113 \\
            & SsGAN (sBN) &     24.355 \\
    \bottomrule
    \end{tabular}
    \captionof{table}{\label{tab:best_fid} Best FID attained across three random seeds.}
\end{minipage}
\end{minipage}

\begin{figure}[ht]
  \centering
  \begin{subfigure}[b]{0.48\textwidth}
    \includegraphics[width=1.0\textwidth]{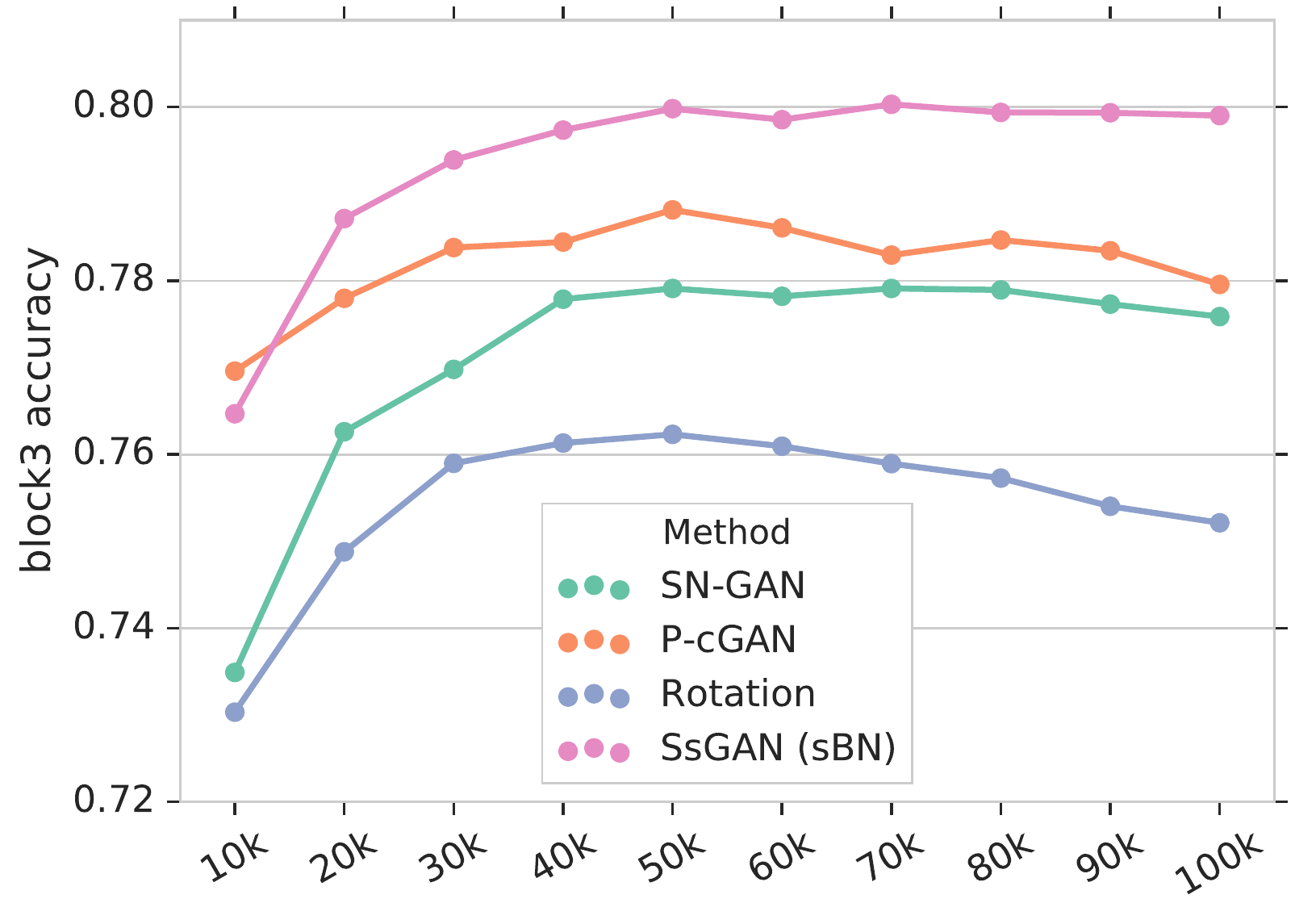}
    \caption{\label{fig:accuracy-steps-cifar-block3}CIFAR-10}
  \end{subfigure}
  \begin{subfigure}[b]{0.48\textwidth}
    \includegraphics[width=1.0\textwidth]{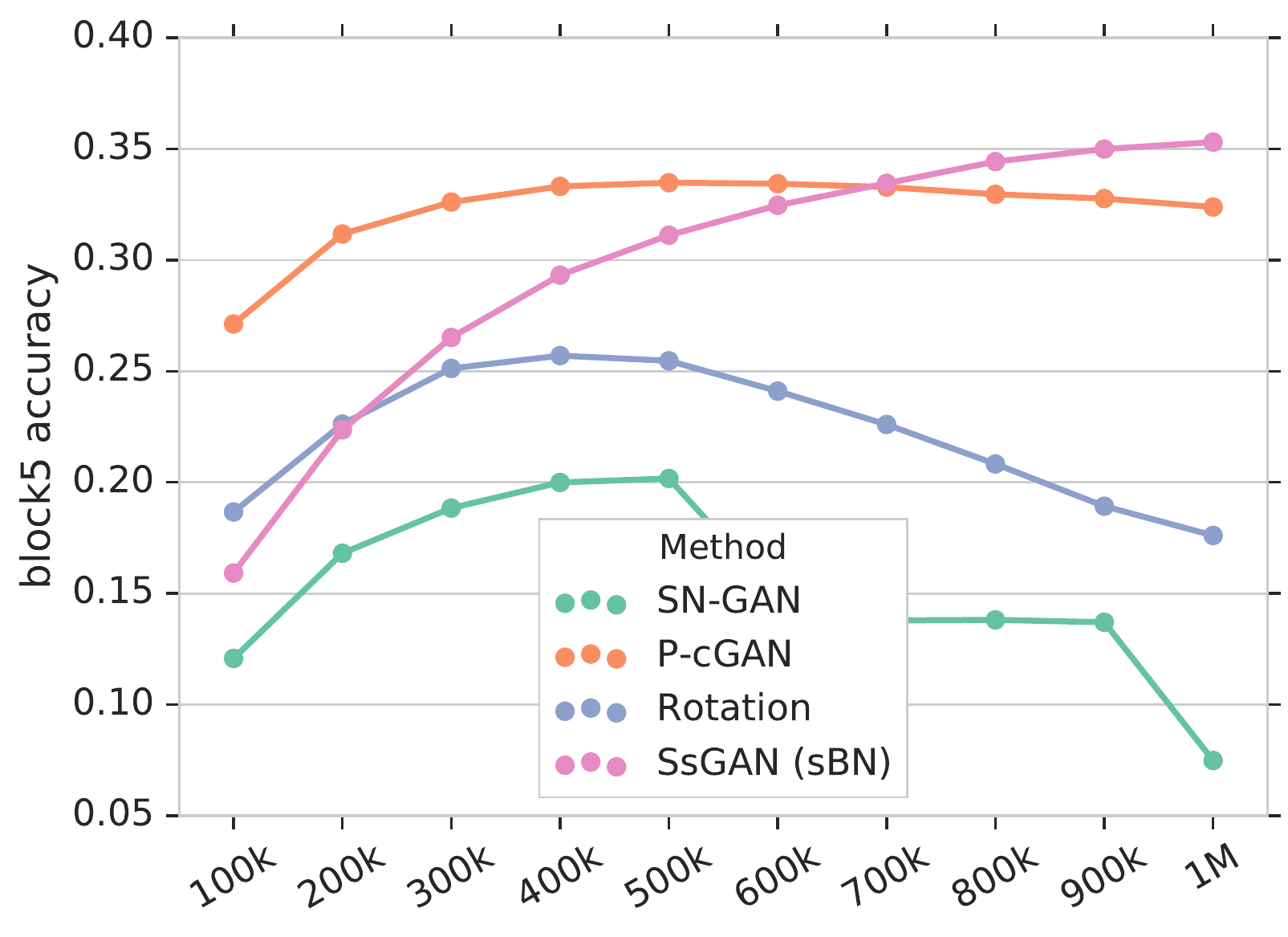}
    \caption{\label{fig:accuracy-steps-imagenet-block5}IMAGENET}
  \end{subfigure}\hfill
  \caption{\label{fig:accuracy-steps}
  Top 1 accuracy (mean across three seeds) to predict labels from discriminator representations from the final ResNet block. 
  x-axis gives the number of GAN training iterations.}
\end{figure}

Table~\ref{tab:best_fid} shows FID of the best run across three random seeds for each dataset/model combination.
Again, for \textsc{cifar} the SsGAN (with sBN) improves unconditional performance, and even equals P-cGAN.
For \textsc{bedroom}, we also see some improvement.
Self supervision appears not to help in \textsc{celeba-hq},
we believe that this data is too simple and face orientations do not provide useful signal.

\paragraph{Representation Quality}
We compare each model's discriminator representations.
We also ablate the GAN loss from SsGAN to assess the representation learned from self-supervision alone.
We followed the evaluation protocol of \cite{zhang2016colorful}. 
We train a logistic regression to predict the class labels from the feature maps of the discriminator
and report top-1 classification accuracy. 

Figure~\ref{fig:accuracy-steps} shows the classification performance using the final ResNet block as a function of GAN training iterations.
SsGAN produces better representations than SN-GAN.
The same is true for all the other blocks (see supplementary, Figures~\ref{fig:accuracy-steps-cifar} and~\ref{fig:accuracy-steps-imagenet}).
Interestingly, during training the SsGAN overtakes P-cGAN, which sees the class labels. This indicates that P-cGAN is overfitting the training data.
Using rotation loss alone substantially decreases representation quality of the SsGAN.
Note that our model uses hyperparameters designed for image generation;
therefore, we do not seek state-of-the-art representations.
Instead, we use representation evaluation to understand and motivate our strategy.
Indeed it seems that self-supervision improves both discriminator representations and image generation.

%% file: content/related.tex




%% file: content/app_model.tex
\section{Additional Details of The Proposed Method}

\begin{figure}[h]
\includegraphics[width=\textwidth]{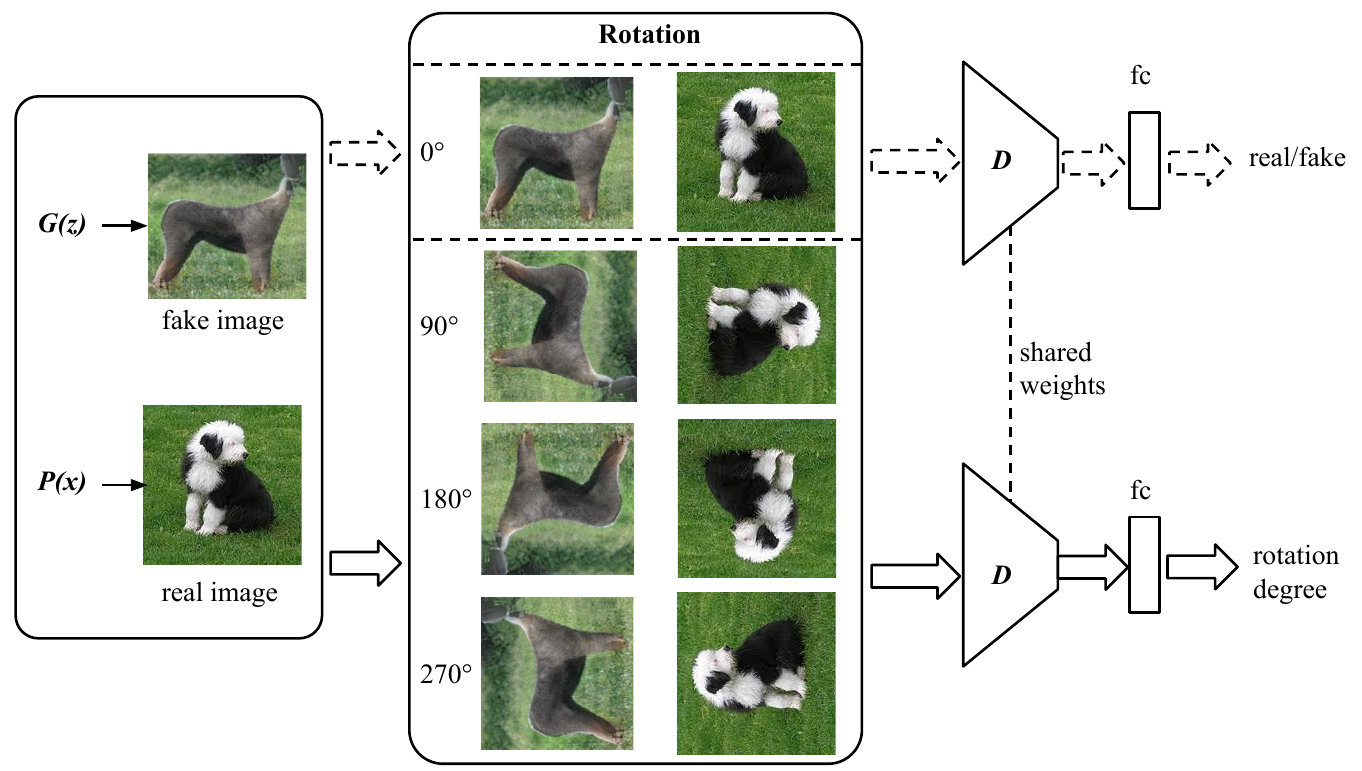}
\caption{\label{fig:rotation_acgan} Rotation feature learning discriminator. The dotted line arrow uses only the non-rotated images, for real/fake classification task. The solid line arrow uses all the four rotated images, for the rotation classification task. }
\end{figure}

Figure~\ref{fig:rotation_acgan} depicts the proposed method. Real and fake/generated images are all rotated four directions. For the shared discriminator D, its goal is to predict true/fake labels of the \textit{non-rotated} images, and detect the rotation of the \textit{real} images.
For the generator, its goal is to generate images to fool discriminator, at the same time trying to generate images that are easier for discriminator to detect rotations (adjusted by $\alpha$).


%% file: content/app_exp.tex
\section{Further Experiments and Details}

\paragraph{Datasets}
We consider four datasets: \cifar{}, \celebahq{}, \lsun{}, and
\imagenet{}.  The \lsun{} dataset~\citep{yu15lsun} contains around 3M
images.  We partition the images randomly into a test set containing
30588 images and a train set containing the rest.  \celebahq{}
contains 30k images~\citep{karras2017progressive}.  We use the
$128\times128\times3$ version obtained by running the code provided by
the authors\footnote{Available at
\url{https://github.com/tkarras/progressive\_growing\_of\_gans}.}.  We
use 3000 examples as the test set and the remaining examples as the
training set.  \cifar{} contains 70K images ($32\times32\times3$),
partitioned into 60000 training instances and 10000 testing instances.
Finally, we evaluate our method on \imagenet{}, which contains
$1.3$M training images and $50$K test images.  We re-size the images
to $128\times128\times3$ as done in~\citet{miyato2018cgans}
and~\citet{zhang2018self}.

\paragraph{Metrics}
To quantitatively evaluate generated samples from different methods, we mainly \emph{Frechet Inception Distance} (FID)~\citep{heusel2017gans}. In FID, the true data and generated samples are first embedded in some feature space (a specific layer of Inception).
Then, a multivariate Gaussian is fit to the data and the distance computed as 
$\FID(x, g) = ||\mu_x -\mu_g||_2^2 + \Tr(\Sigma_x + \Sigma_g - 2(\Sigma_x\Sigma_g)^\frac12)$, 
where
$\mu$ and $\Sigma$ denote the empirical mean and covariance
and subscripts $x$ and $g$ denote the true and generated data respectively.
As shown in \citet{sajjadi2018assessing,lucic2018}, FID is shown to be sensitive to both the addition of spurious modes and to mode dropping. 

\paragraph{Additional experimental details}
When the rotation loss is added, we take a quarter of images in the batch, rotate for all four directions to synthetic images of known rotation orientations. This subset rotation trick could reduce and keep the extra computation small. It is worth noting that other than the extra loss term, other parts (architectures, hyper-parameters) are all kept the same for our method and baselines (both unconditional and conditional GANs). For simplicity, we fix $\beta=1$ and compare $\alpha$ in $\{0.2, 0.5, 1\}$; $\alpha=0.2$ is used by default.

We use the Adam optimizer. The learning rate is fixed for both datastes as $0.0002$. All models are trained for $100$k steps\footnote{Results in the main paper for \imagenet{} are obtained from training with 1M steps.}. All models are trained with batch size of 64 in a single Nvidia P100 GPU. Reported results by default are the mean of from 3 independent runs with different random seeds. 

\paragraph{Robustness test}
We test the effects of training stabilization brought by the SsGAN. 
We consider two types of hyper-parameter settings: First, controlling the Lipschitz constant of the discriminator, which is a central quantity analyzed in the GAN literature~\citep{ miyato2018spectral,zhou2018understanding}. 
We
consider two state-of-the-art techniques: Gradient Penalty~\citep{gulrajani2017improved}, and Spectral Normalization~\citep{miyato2018spectral}. 
For the Gradient Penalty regularizer we consider regularization strength $\lambda \in \{1,10\}$. 
Second, we consider different Adam optimizer hyper-parameters.
We test two popular settings $(\beta_1, \beta_2)$: $(0.5, 0.999)$ and $(0,0.9)$.  Previous studies find that multiple discriminator steps per generator step help training~\citep{goodfellow2014generative,salimans2016improved}, thus we also consider both $1$ and $2$ discriminator steps per generator step\footnote{We also experimented with 5 steps which didn't outperform the $2$ step setting.}.  In total, this amounts to three different sets of hyper-parameters for $(\beta_1, \beta_2,
\text{disc\_iter})$: $(0, 0.9, 1)$, $(0, 0.9, 2)$, $(0.5, 0.999, 1)$.


\subsection{Quantitative Comparisons of the Generative Samples}
\label{fid_quantitative_compare}



\begin{figure}[ht]
  \centering
  \begin{subfigure}[b]{0.425\textwidth}
    \includegraphics[width=1\textwidth]{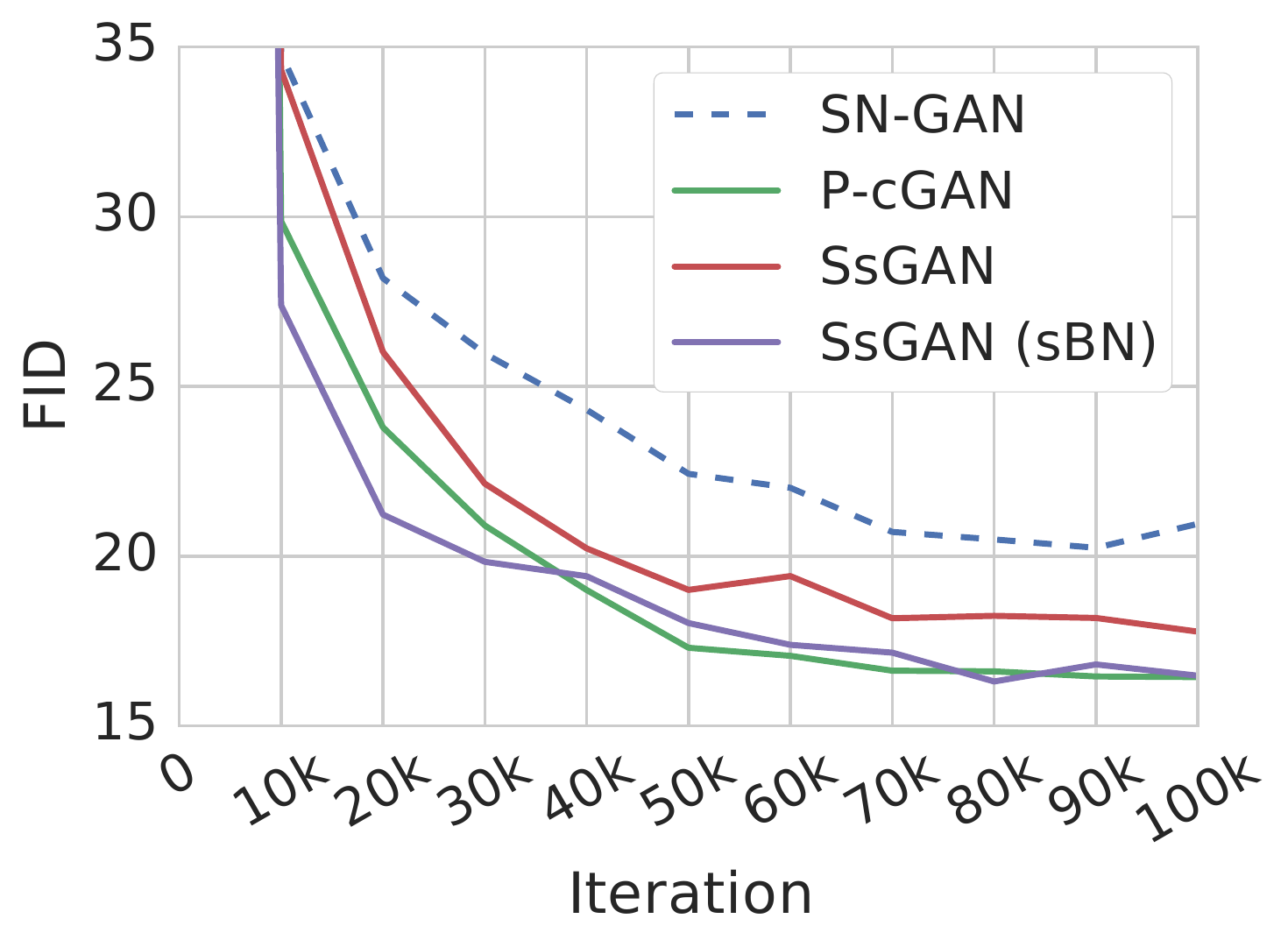}
    \caption{\cifar{}}
  \end{subfigure}
  \begin{subfigure}[b]{0.4\textwidth}
    \includegraphics[width=1\textwidth]{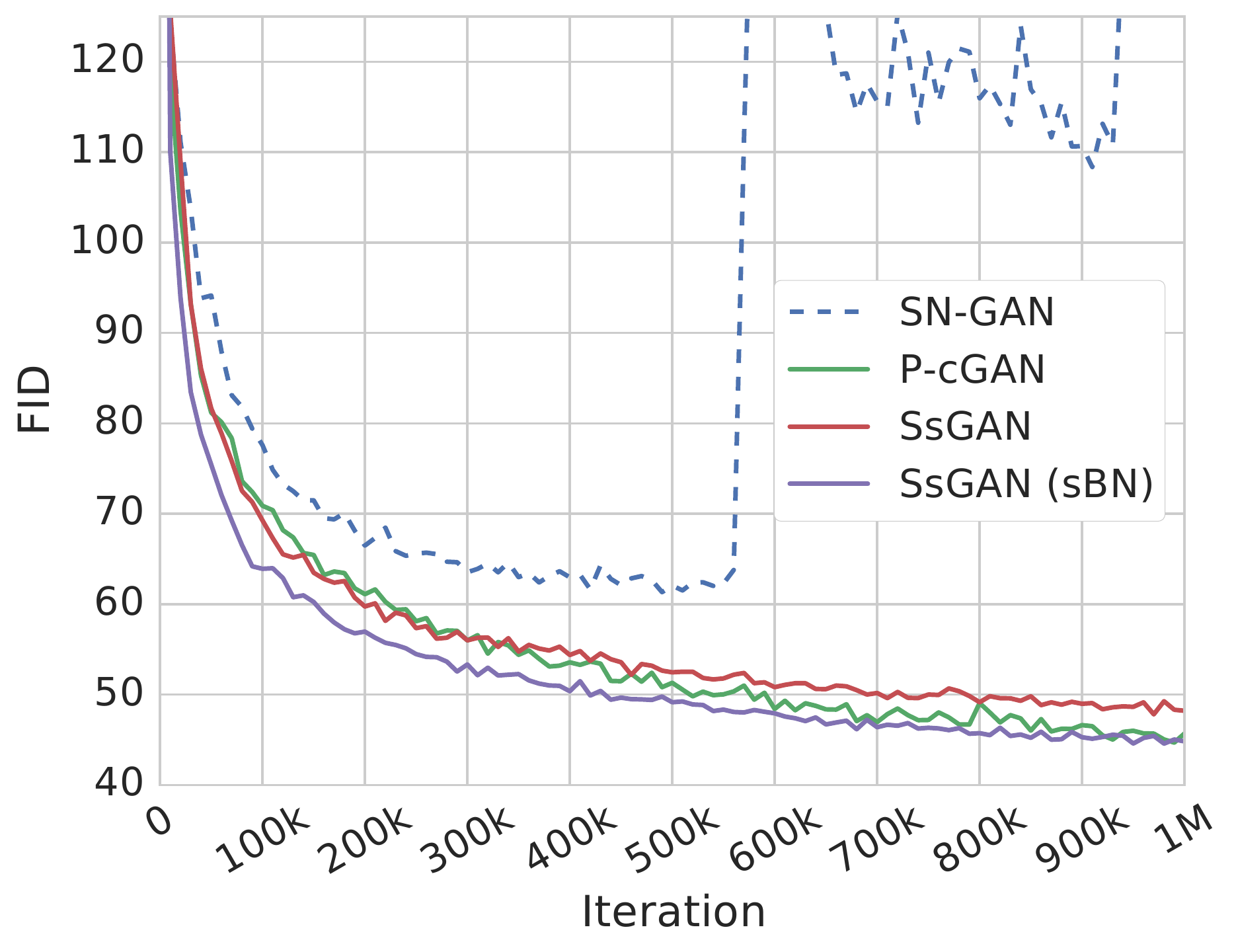}
    \caption{\imagenet{}}
  \end{subfigure}\hfill
  \caption{\label{fig:convergence_cruves} FID convergence curves for different methods with 100k training steps. The self supervision close the gap between the conditional and unconditional GANs, sometimes even surpass the class label conditional GAN.}
\end{figure}

\paragraph{Convergence curve} Figure \ref{fig:convergence_cruves} shows the convergence curves of different methods. We can clearly see the proposed SsGAN significantly improves the performance over the other unconditional GAN, SN-GAN, and even matches the conditional P-cGAN.

\paragraph{Robustness improvement} The use of self-supervised loss not only improves the performance of the default hyper-parameters we use, but also improves under multiple types of hyper-parameters. Table \ref{tab:gp_robustness_fid} shows the performance comparison with Gradient Penalty with various hyper-parameters, and Table \ref{tab:sn_robustness_fid} shows the performance comparison with Spectral Normalization with various hyper-parameters. Both figures demonstrate SsGAN reduces the sensitivity of GAN training to hyper-parameters, hence the training is more robust.

\begin{table}
\centering
\small
\caption{\label{tab:gp_robustness_fid} FID comparisons for unconditional GANs under various hyper-parameter settings with gradient penalty. Self-supervision loss reduces the sensitivity of GAN training to hyper-parameters.}
\begin{tabular}{cccccccc}
\toprule
   &     &       &  & \multicolumn{2}{c}{\cifar{}} & \multicolumn{2}{c}{\imagenet{}} \\
$\lambda$ & $\beta_1$ & $\beta_2$ & Disc\_iters  &      SN-GAN &   SsGAN &      SN-GAN &    SsGAN \\
\midrule
\multirow{3}{*}{1} & \multirow{2}{*}{0.0} & \multirow{2}{*}{0.900} & 1 &  121.047 &  \textbf{25.798} &  183.364 &   \textbf{80.669} \\ \cline{4-8}
   &     &       & 2 &   28.107 &  \textbf{26.984} &   85.130 &   \textbf{83.081} \\
\cline{2-8}
\cline{3-8}
   & 0.5 & 0.999 & 1 &   78.542 &  \textbf{25.890} &  104.731 &   \textbf{91.628} \\
\cline{1-8}
\multirow{3}{*}{10} & \multirow{2}{*}{0.0} & \multirow{2}{*}{0.900} & 1 &  188.523 &  \textbf{28.481} &  227.041 &   \textbf{85.376} \\ \cline{4-8}
   &     &       & 2 &   29.110 &  \textbf{27.735} &  227.738 &   \textbf{80.822} \\
\cline{2-8}
\cline{3-8}
   & 0.5 & 0.999 & 1 &  117.671 &  \textbf{25.216} &  242.705 &  \textbf{144.353} \\
\bottomrule
\end{tabular}
\end{table}

\begin{table}
\centering
\small
\caption{\label{tab:sn_robustness_fid} FID comparisons for unconditional GANs under various hyper-parameter settings with spectral normalization. Self-supervision loss reduces the sensitivity of GAN training to hyper-parameters.}
\begin{tabular}{cccccccc}
\toprule
    &       &  & \multicolumn{2}{c}{\cifar{}} & \multicolumn{2}{c}{\imagenet{}} \\
$\beta_1$ & $\beta_2$ & Disc\_iters  &     SN-GAN &   SsGAN &      SN-GAN &   SsGAN \\
\midrule
\multirow{2}{*}{0.0} & \multirow{2}{*}{0.900} & 1 &  87.864 &  \textbf{19.645} &  129.964 &  \textbf{86.095} \\ \cline{3-7}
    &       & 2 &  20.245 &  \textbf{17.885} &   80.047 &  \textbf{70.637} \\
\cline{1-7}
\cline{2-7}
0.5 & 0.999 & 1 &  86.869 &  \textbf{18.235} &  201.936 &  \textbf{99.968} \\
\bottomrule
\end{tabular}
\end{table}

\begin{figure}[h]
  \centering
  \begin{subfigure}[b]{0.26\textwidth}
    \includegraphics[width=1\textwidth]{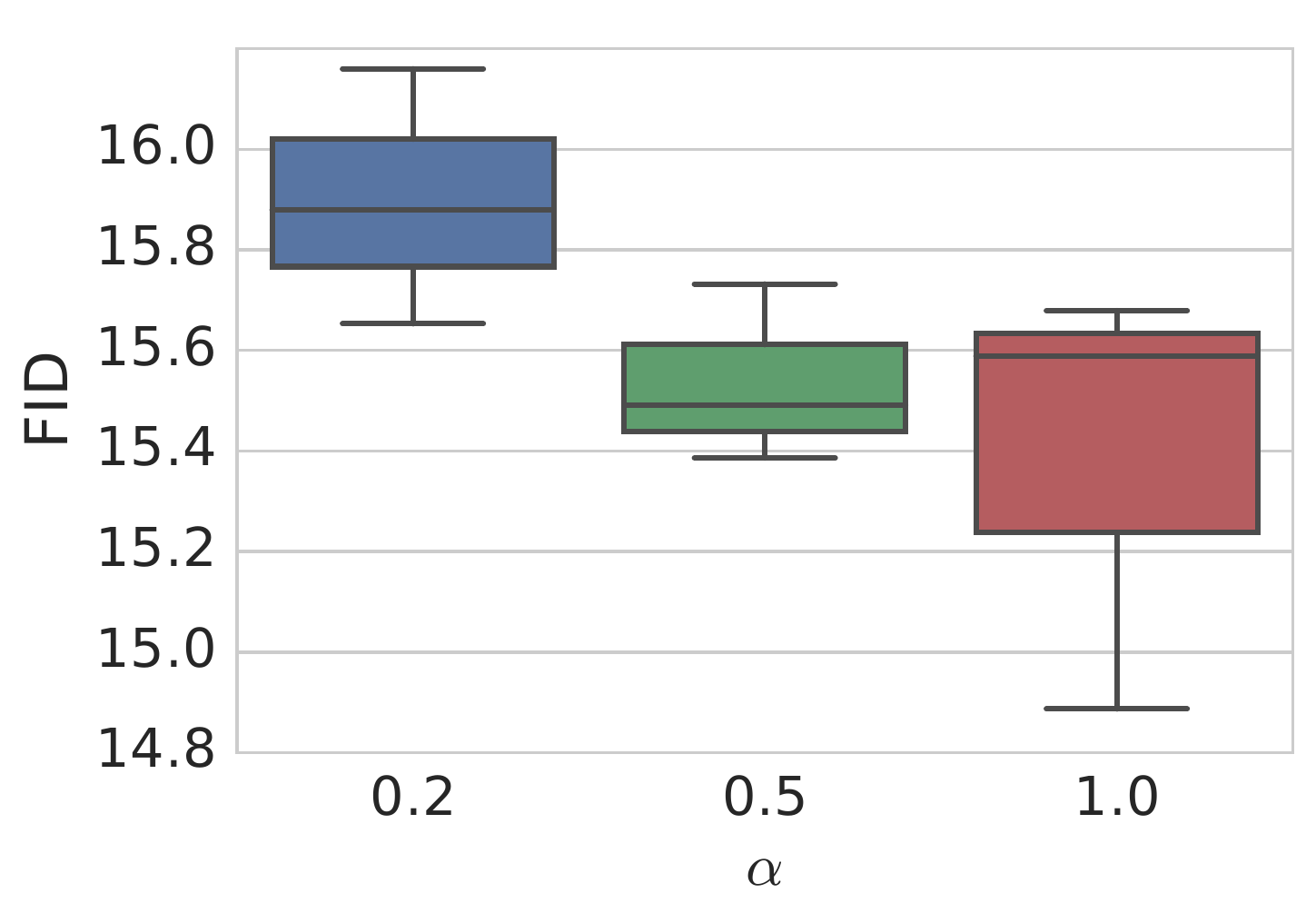}
    \caption{\cifar{}}
  \end{subfigure}
  \begin{subfigure}[b]{0.25\textwidth}
    \includegraphics[width=1\textwidth]{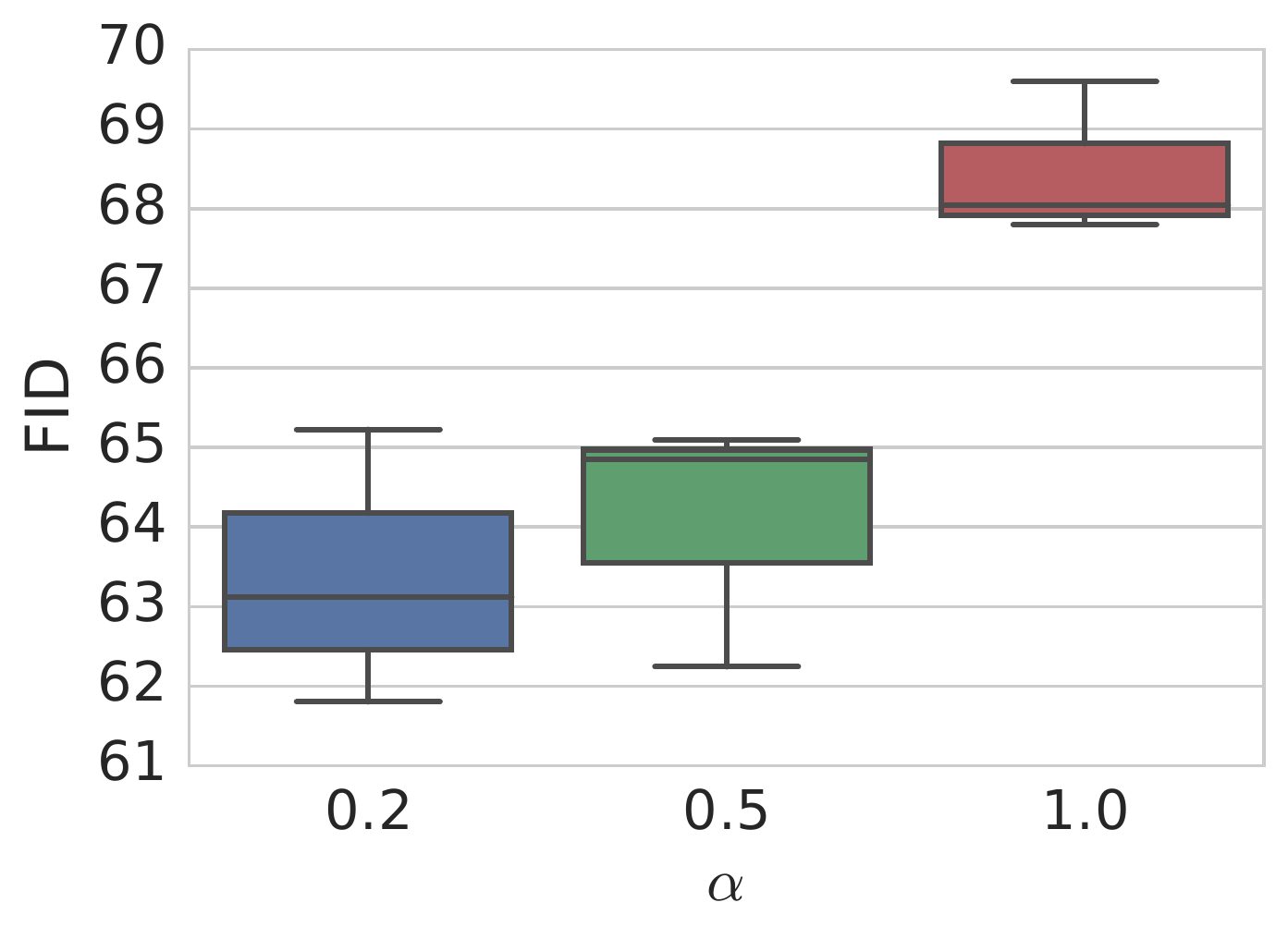}
    \caption{\imagenet{}}
  \end{subfigure}
  \caption{\label{fig:beta_variant} Performance under different $\alpha$ values.}
\end{figure}

\paragraph{Selection of the hyper-parameters} We further compare different choices of $\alpha$ (while fixing $\beta=1$). As discussed, an inappropriate value of $\alpha$ could bias the generator, while a reasonable one would help. As shown in Table \ref{fig:beta_variant}, we indeed see the effectiveness of $\alpha$. In the values compared, the optimal $\alpha$ is 1 for \cifar{}, and 0.2 for \imagenet{}. By default we use 0.2 for all datasets for simplicity.

\subsection{Qualitative Comparisons of the Generative Samples}

\begin{figure}[ht]

\begin{subfigure}[b]{0.49\textwidth}
    \begin{center}
     \includegraphics[width=1\textwidth]{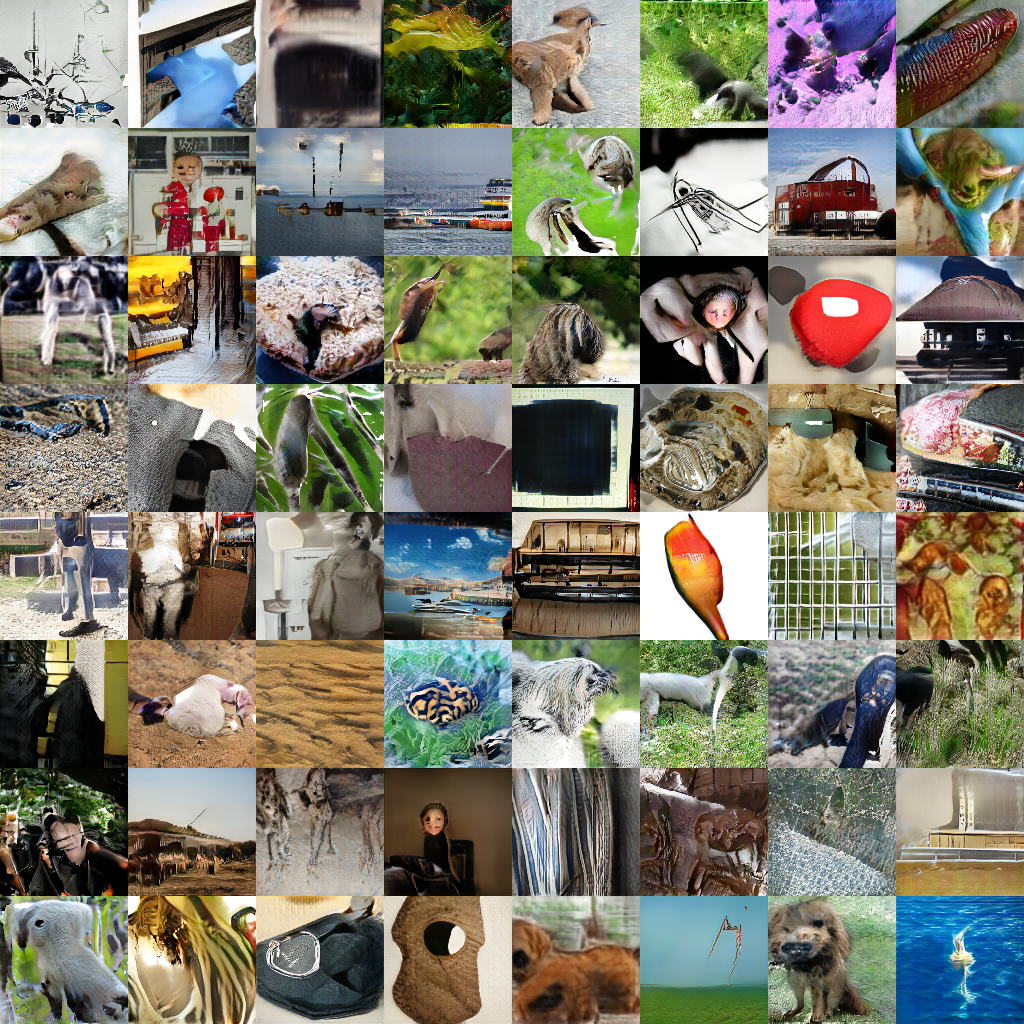}
    \end{center}
    \caption{SN-GAN (FID=59.3)}
\end{subfigure}\hfill
\begin{subfigure}[b]{0.49\textwidth}
    \begin{center}
     \includegraphics[width=1\textwidth]{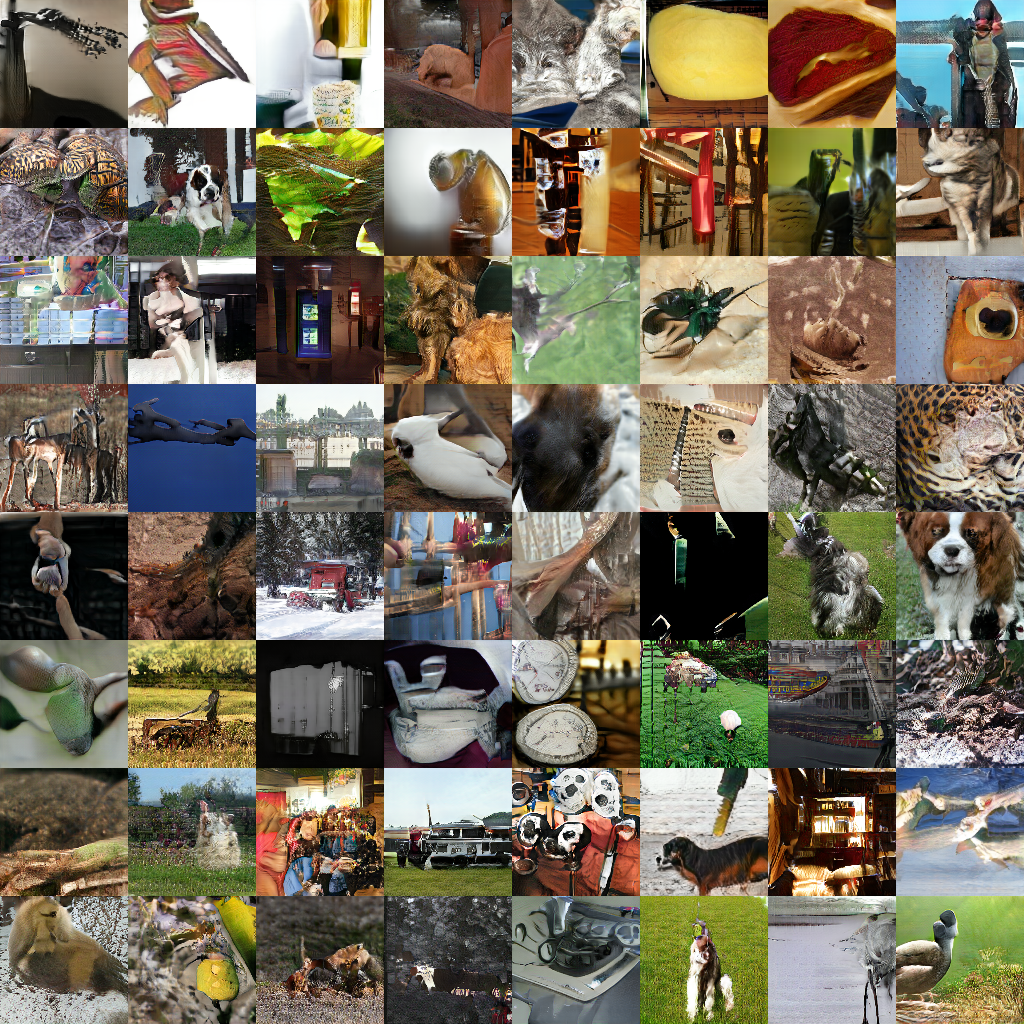}
    \end{center}
    \caption{P-cGAN (FID=43.0)}
\end{subfigure}\hfill
\begin{subfigure}[b]{0.49\textwidth}
    \begin{center}
     \includegraphics[width=1\textwidth]{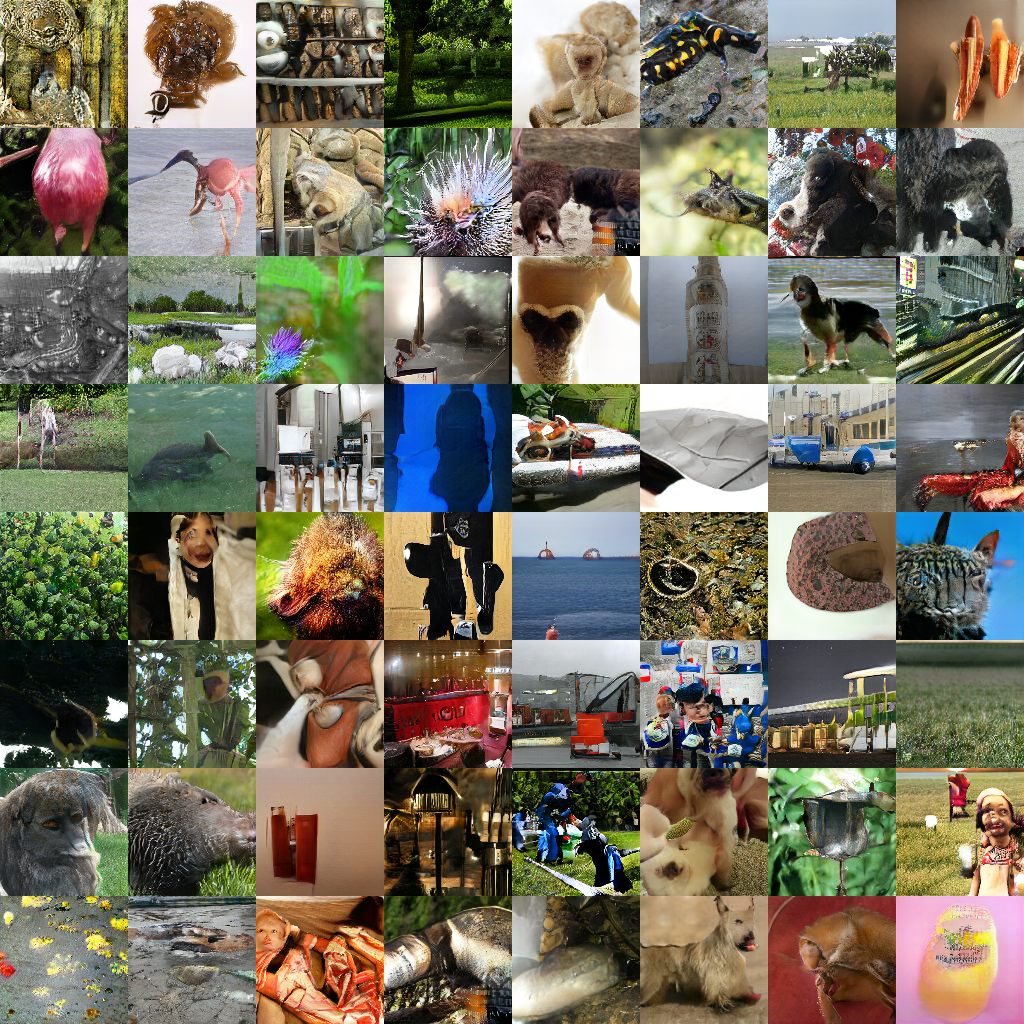}
    \end{center}
    \caption{SsGAN (FID=47.8)}
\end{subfigure}\hfill
\begin{subfigure}[b]{0.49\textwidth}
    \begin{center}
     \includegraphics[width=1\textwidth]{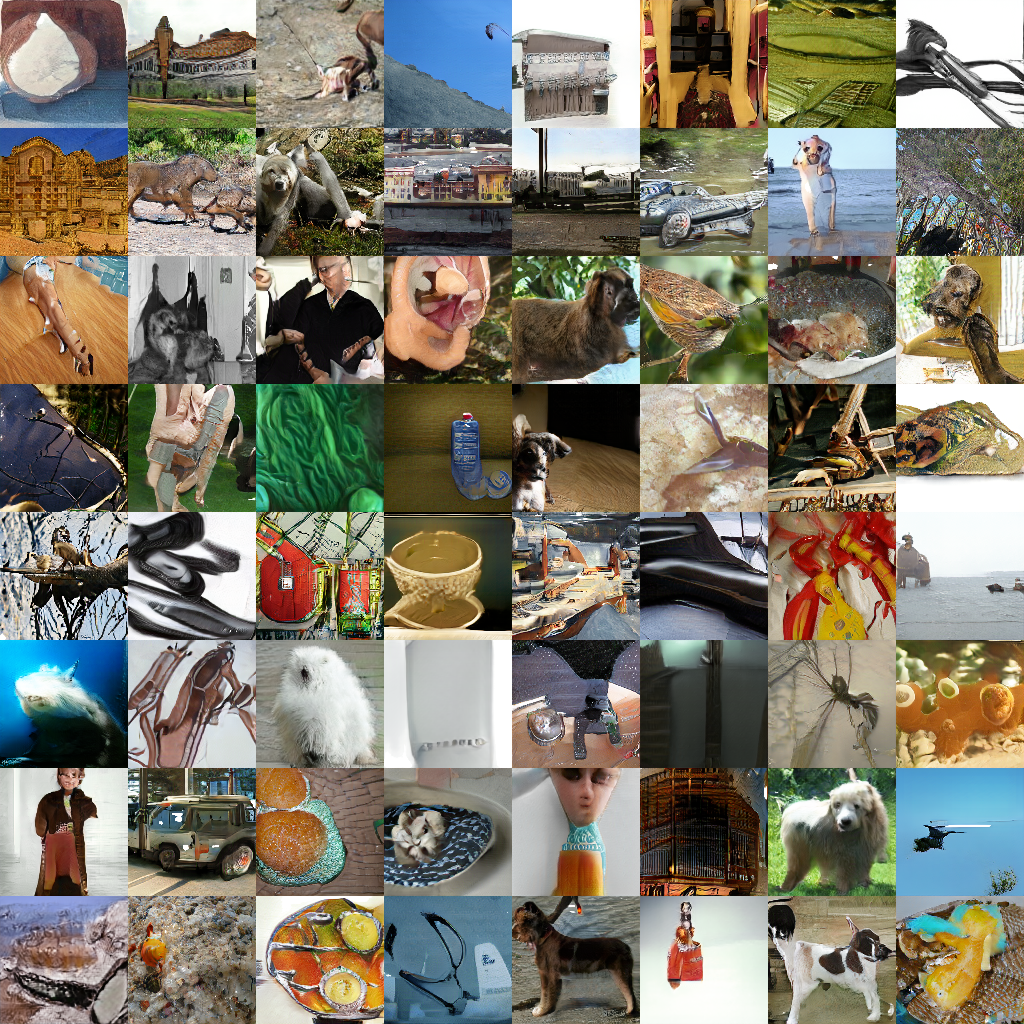}
    \end{center}
    \caption{SsGAN (sBN) (FID=43.9)}
\end{subfigure}\hfill
\caption{\label{fig:samples_imagenet} Random samples generated by GANs described in this paper on \textsc{imagenet} (at 1M steps).}
\end{figure}

Figure \ref{fig:samples_imagenet} shows the generated examples from different GANs for \textsc{imagenet}. Although the baseline unconditional GAN generates sharp images, majority of the generated objects e.g. dogs are distorted. With the supervised labels, P-cGAN's samples are improved, e.g. we see the dog faces and diverse objects. SsGAN generates similar quality images to the conditional P-cGAN. 


\subsection{Comparison of Representation Quality}

We compare the representation qualities of the discriminator's intermediate layers. The comparison is performed by training logistic regression classifiers on top of the feature maps from each ResNet block to perform the 1000-way \imagenet{} classification task, as proposed in \citet{zhang2016colorful}. The representation features are spatially resized 
to around 9000 dimension with adaptive max pooling. We report top-1 classification accuracy. 

We compare the SN-GAN, P-cGAN and SsGAN (SBN) models with the best FID score from Section \ref{fid_quantitative_compare}. We also consider ablating the GAN loss from SsGAN, to yield a rotation only model (Rotation). Rotation uses the same architecture and hyper parameters as the SsGAN discriminator.
Evaluation is performed on three independent models with different random seeds, where the mean accuracy and the standard deviation are reported. We tune the learning rate for linear regression on a validation set created by holding out $10\%$ of the training set. The linear regression results are insensitive to the hyperparameters. On \cifar{}, we use SGD with batch size 256, momentum 0.9 and learning rate 0.1. We drop the learning rates by a factor of 10 after every 30 epochs. We train in total for 100 epochs, where the learning rates are dropped by a factor of 10 after every 30 epochs. 
On \imagenet{}, we use SGD with batch size 64, momentum 0.9 and learning rate 0.01. We train in total for 30 epochs, where the learning rates are dropped by a factor of 10 after every 10 epochs. 

\begin{figure}[h]
\begin{center}
\epsfig{file=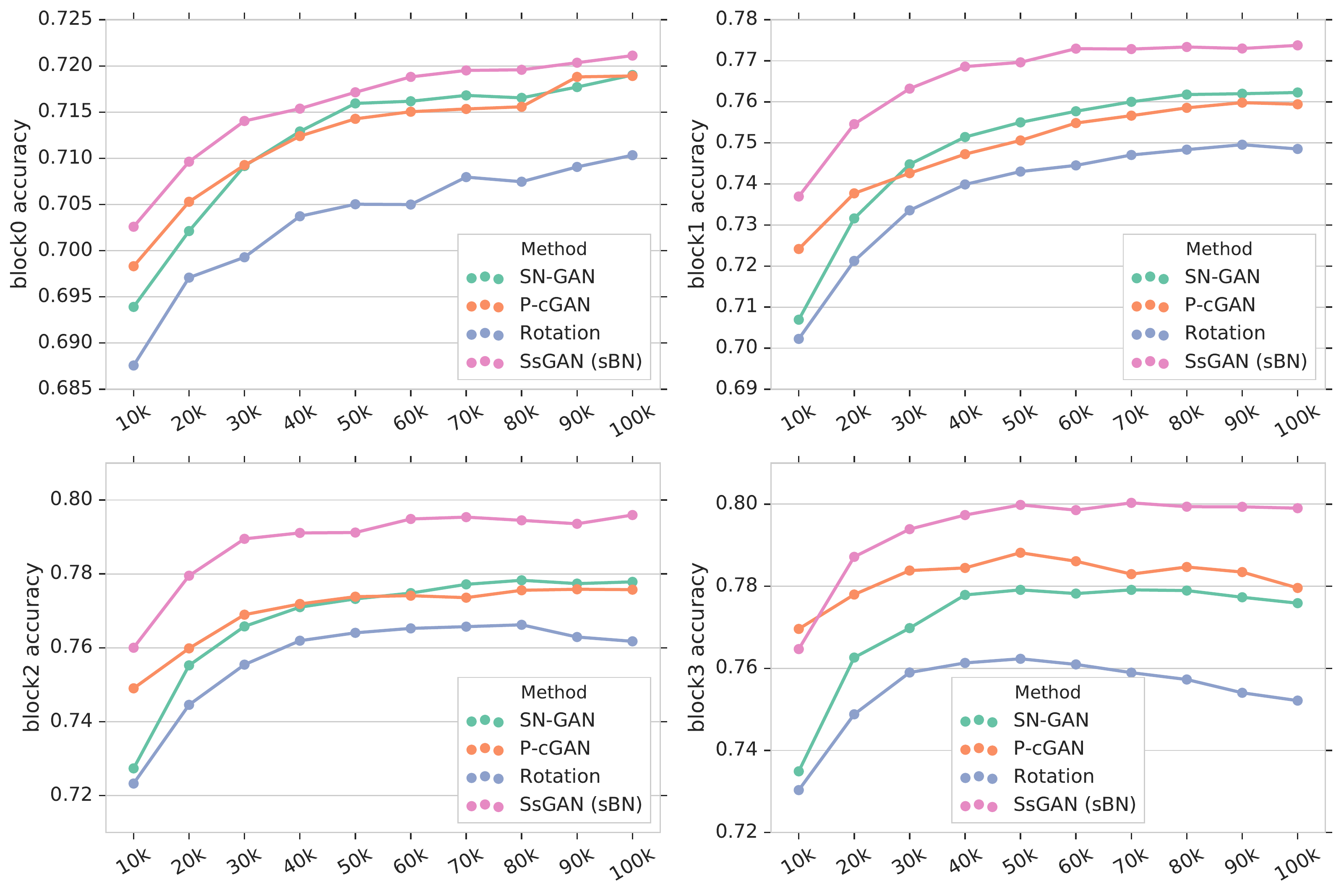,height=8.0cm}
\end{center}
\caption{\label{fig:accuracy-steps-cifar} Top 1 accuracy on \cifar{} with training steps from 10k to 100k.}
\end{figure}

Table~\ref{tab:cifar10_top1} shows the top-1 classification accuracy on \cifar{}. Figure~\ref{fig:accuracy-steps-cifar} shows the results as s function of training steps of the original model. We plot the quality of representations extracted from each of the the 4 blocks for ResNet \cifar architecture. The conditional model, P-cGAN, produces  similar results to the other unsupervised model GAN and Rotation model. The SsGAN outperforms the other models under most conditions. 

\begin{figure}[h]
\begin{center}
\epsfig{file=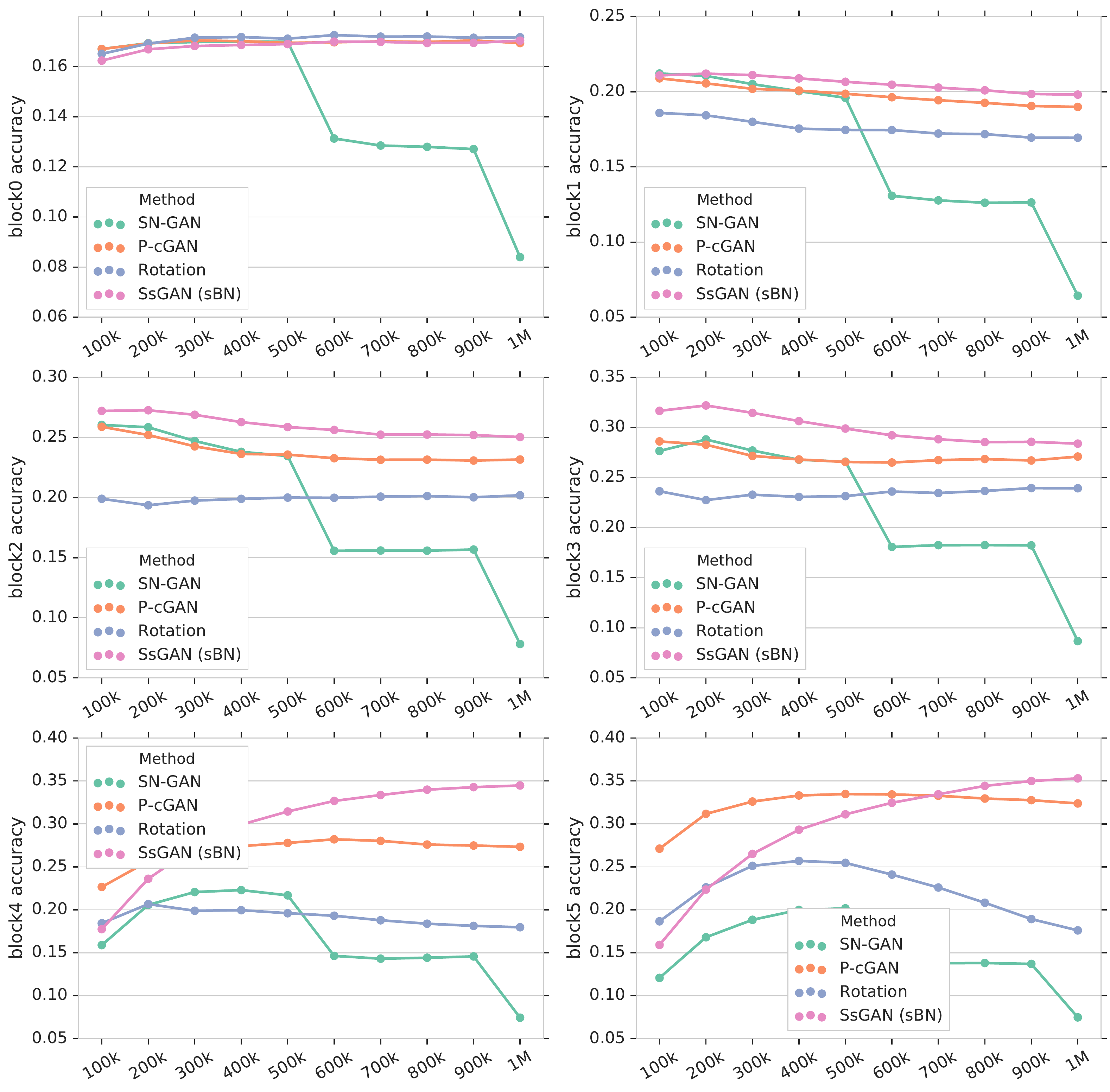,height=12.0cm}
\end{center}
\caption{\label{fig:accuracy-steps-imagenet} Top 1 accuracy on \imagenet{} validation set with training steps from 10k to 1M.
}
\end{figure}

Table \ref{tab:imagenet_top1} and Figure~\ref{fig:accuracy-steps-imagenet} show results on \imagenet{}. 
There are six blocks in the \imagenet{} Resnet architecture. 
For SN-GAN and Rotation, block3 performs best, for P-cGAN and SsGAN, block 5 performs best. 
It is expected that P-cGAN would benefit from representations closer to the classification layer because it sees the actual labels being used in the downstream representation evaluation.
In block5, we observe that SsGAN and P-cGAN representation quality is not improving after 200K steps, while SsGAN improves the representation quality for all 1M training steps.
Overall, representations are improved with self-supervised GAN and conditional GAN, which correlated with their improved samples. 

Table~\ref{tab:imagenet_compare} shows the comparison of our method to other self supervised learning methods. We select the best run according to the validation results. Overall, the SsGAN model achieves comparable results on \imagenet{}. Note that our method is not tuned for representation, but rather for image generation. Among those methods, BiGAN~\citep{donahue2016adversarial} learns the representation from GANs as well. BiGAN learns the representation with an additional encoder network, while our method is arguably simpler because it extracts the representation directly from the discriminator network.

\begin{table}[h]
\centering
\caption{\label{tab:cifar10_top1} Top-1 accuracy on \cifar.}

\begin{tabular}{llllll}
\toprule
Method & SN-GAN & P-cGAN & Rotation & SsGAN (sBN) \\
\midrule
Block0    &   $0.719 \pm 0.002$ &   $0.719 \pm 0.003$ & $0.710 \pm 0.002$ & $0.721 \pm 0.002$\\
Block1      &   $0.762 \pm  0.001$ &   $0.759 \pm  0.003$ & $0.749 \pm 0.003$ & $0.774 \pm 0.003$ \\
Block2 &   $\mathbf{0.778 \pm  0.001}$ &   $0.776 \pm  0.005$ & $\mathbf{0.762 \pm 0.003}$ & $0.796 \pm 0.005$ \\
Block3  &  $0.776 \pm  0.005$ &   $\mathbf{0.780 \pm  0.006}$ & $0.752 \pm 0.006$ & $\mathbf{0.799 \pm 0.003}$ \\
\bottomrule
\end{tabular}
\end{table}

\begin{table}[h]
\centering
\small
\caption{\label{tab:imagenet_top1} Top-1 accuracy on \imagenet{} for all Resnet blocks.}

\begin{tabular}{lllll}
\toprule
Method & SN-GAN & P-cGAN & Rotation & SsGAN (sBN) \\
\midrule
Block0      &    $0.084 \pm 0.076$   &    $0.169 \pm  0.001$    &    $0.172 \pm  0.002$    &   $0.170 \pm  0.001$ \\
Block1      &    $0.064 \pm  0.107$   &    $0.190 \pm  0.004$    &    $0.169 \pm  0.005$    &   $0.198 \pm  0.003$ \\
Block2      &    $0.078 \pm  0.134$   &    $0.232 \pm 0.002 $    &    $0.202 \pm 0.004$     &   $0.250 \pm 0.001$ \\
Block3      &    $\mathbf{0.087 \pm 0.149} $   &    $0.271 \pm 0.003$    &    $\mathbf{0.239 \pm 0.005}$     &   $0.284 \pm 0.002$ \\
Block4      &    $0.0744 \pm 0.127$   &    $0.273 \pm 0.019  $    &    $0.180 \pm 0.003$     &   $0.345 \pm 0.003$ \\
Block5      &    $0.075 \pm 0.128$   &    $\mathbf{0.324 \pm 0.007}$    &    $0.176 \pm 0.033$     &   $\mathbf{0.353 \pm 0.006}$ \\
\bottomrule
\end{tabular}
\end{table}

\begin{table}[h]
\centering
\small
\caption{\label{tab:imagenet_compare} Comparison with other self-supervised representation learning methods by top-1 accuracy on \imagenet{}.}

\begin{tabular}{llllll}
\toprule
Method & Context~\citep{doersch2015unsupervised} & BiGAN~\citep{donahue2016adversarial} & Colorization~\citep{zhang2016colorful} & RotNet~\citep{gidaris2018unsupervised} & SsGAN (sBN) \\
\midrule
Accuracy      &    0.317   &    0.310    &    0.315    &   0.387 & 0.359 \\
\bottomrule
\end{tabular}
\end{table}

